\documentclass[11pt]{article}
\usepackage{coling2020}
\usepackage{times}
\usepackage{url}
\usepackage{latexsym}
\usepackage{graphicx}
\usepackage{subcaption}
\usepackage{todonotes}
\usepackage{amssymb}

\usepackage{xcolor}
\colingfinalcopy 

\title{Exploring Cross-sentence Contexts for\\Named Entity Recognition with BERT}

\author{Jouni Luoma \\
  TurkuNLP group \\
  University of Turku \\
  Turku, Finland \\
  {\tt jouni.a.luoma@utu.fi} \\\And
  Sampo Pyysalo \\
  TurkuNLP group \\
  University of Turku \\
  Turku, Finland \\
  {\tt sampo.pyysalo@utu.fi} \\}

\date{}

\begin{document}
\maketitle
\begin{abstract}
Named entity recognition (NER) is frequently addressed as a sequence classification task with each input consisting of one sentence of text. It is nevertheless clear that useful information for NER is often found also elsewhere in text. Recent self-attention models like BERT can both capture long-distance relationships in input and represent inputs consisting of several sentences. This creates opportunities for adding cross-sentence information in natural language processing tasks. This paper presents a systematic study exploring the use of cross-sentence information for NER using BERT models in five languages. We find that adding context as additional sentences to BERT input systematically increases NER performance. Multiple sentences in input samples allows us to study the predictions of the sentences in different contexts. We propose a straightforward method, Contextual Majority Voting (CMV), to combine these different predictions and demonstrate this to further increase NER performance. Evaluation on established datasets, including the CoNLL'02 and CoNLL'03 NER benchmarks, demonstrates that our proposed approach can improve on the state-of-the-art NER results on English, Dutch, and Finnish, achieves the best reported BERT-based results on German, and is on par with other BERT-based approaches in Spanish. We release all methods implemented in this work under open licenses.

\end{abstract}

\section{Introduction}
\label{intro}

Named entity recognition (NER) approaches have evolved through various methodological phases, broadly including rule/knowledge-based, unsupervised, feature engineering and supervised learning, and feature inferring approaches \cite{yadav-bethard-2018-survey,Li_2020}. The use of cross-sentence information in some form has been a normal part of many NER methods in the former categories, but its role has diminished with the current feature inferring deep learning based approaches. Rule/knowledge-based approaches such as that of \newcite{mikheev-etal-1998-description} typically match strings to lexicons and similar domain knowledge sources, possibly going through text multiple times with refinement based on entities found on earlier passes. Later, manually engineered features were used to incorporate information from the surrounding text, whole documents, data sets and also from external sources. The number of different features and classifiers grew during the years and it was normal that the features also contained cross-sentence information in some form as for example in \cite{krishnan-manning-2006-effective}. Dense representations of text such as word, character, string and subword embeddings first started to appear in NER methods as additional features given to classifiers \cite{collobert2011}. Step by step, feature engineering has been demoted to a lesser role, as the most recent deep learning approaches learn to create meaningful and context-sensitive representations of text by pre-training with vast amounts of unlabelled data. These contextual representations are often used directly as features for existing NER architectures or fine-tuned with labelled data to match a certain task. 

%
%
\blfootnote{
    %
    %
    %
    
    \hspace{-0.65cm}  
    This work is licensed under a Creative Commons 
    Attribution 4.0 International Licence.
    Licence details:
    \url{http://creativecommons.org/licenses/by/4.0/}.
     
    %
}

In recent years, the development of Natural Language Processing (NLP) in general and NER in particular have been greatly influenced by deep transfer learning methods capable of creating contextual representations of text, to the extent that many of the state-of-the-art NER systems mainly differ from one another on the basis of how these contextual representations are created \cite{Peters_2018,devlin2018bert,akbik2018coling,Baevski_2019}. Using such models, sequence tagging tasks are often approached one sentence at a time, essentially discarding any information available in the broader surrounding context, and there is only little recent study on the use of cross-sentence context -- sentences around the sentence of interest -- to improve sequence tagging performance. In this paper, we present a comprehensive exploration of the use of cross-sentence context for named entity recognition, focusing on the recent BERT deep transfer learning models \cite{devlin2018bert} based on self-attention and the transformer architecture \cite{vaswani2017attention}. BERT uses a fixed-size window that limits the amount of text that can be input to the model at one time. The model maximum window size, or \emph{maximum sequence length}, is fixed during pre-training, with 512 wordpieces a common choice. This window fits dozens of typical sentences of input at a time, allowing us to include extensive sentence context. Here, we first study the effect of predicting tags for individual sentences when they are moved around the window, surrounded by their original document context from the source data. Second, we utilize different predictions for the same sentences to potentially further improve performance, combining predictions using majority voting, adapting an approach that has been used already in early NER implementations \cite{tjong-kim-sang-etal-2000-applying,van-halteren-etal-2001-improving,florian-etal-2003-named}.
We evaluate these approaches on five languages, contrasting NER results using BERT without cross-sentence information, sentences in context, and aggregation using Contextual Majority Voting (CMV) on well-established benchmark datasets.
We show that using sentences in context consistently improves NER results on all of the tested languages and CMV further improves the results in most cases. Comparing performance to the current state-of-the-art NER results in the 5 languages, we find that our approach establishes new state-of-the-art results for English, Dutch, and Finnish, the best BERT-based results on German, and effectively matches the performance of a BERT-based method in Spanish.

\section{Related work}

The state-of-the-art in NER has recently moved from approaches using word/character representations and manually engineered features \cite{passos-etal-2014-lexicon,Chiu_2016} toward approaches directly utilizing deep learning-based contextual representations \cite{akbik2018coling,Peters_2018,devlin2018bert,Baevski_2019} while adding few manually engineered features, if any. While successful in terms of NER performance, these approaches have tended to predict tags for one sentence at a time, discarding information from surrounding sentences.

One recent method taking sentence context into account is that of \newcite{akbik-etal-2019-pooled}, which addresses a weakness of an earlier contextual string embedding method \cite{akbik2018coling}, specifically the issue of rare word representations occurring in underspecified contexts. \newcite{akbik-etal-2019-pooled} make the intuitive assumption that such occurrences happen when a named entity is expected to be known to the reader, i.e.\ the name is either introduced earlier in text or is of general in-domain knowledge. Their approach is to maintain a memory of contextual representations of each unique word/string in text and pool together contextual embeddings of a string occurring in text with the contextual embeddings of the same string earlier in text. This pooled contextual embedding is then concatenated with the current contextual embedding to get the final embedding to use in classification.

Another recent approach taking broader context into account for NER was proposed by \newcite{luo2020hierarchical}, where in addition to token representations, also sentence and document level representations are calculated and used for classification using a CRF model. A sliding window is used by \newcite{Wu_2019} so that part of the input is preserved as context when the window is moved forward in text. 
\newcite{Baevski_2019} state that they use longer paragraphs in pre-training their model, but it is not mentioned in the paper if such longer paragraphs are used also in fine-tuning the model or predicting tags for NER.
Some other approaches such as that of \newcite{liu-etal-2019-towards} include explicit global information in the form of e.g.\ gazetteers. Also, some approaches formulate NER as a span finding task instead of sequence labelling \cite{banerjee2019knowledge,li-etal-2020-unified}. These approaches would likely allow the use of longer sequences, but the incorporation of cross-sentence information is not explicitly proposed by the authors. In the paper introducing BERT, \newcite{devlin2018bert} write in the description of their NER evaluation ``we include the maximal document context provided by the data.'' However, no detailed description of how this inclusion was implemented is provided, and some NER implementations using BERT have struggled to reproduce the results of the paper.\footnote{\url{https://github.com/google-research/bert/issues/581}}\textsuperscript{,}\footnote{\url{ https://github.com/google-research/bert/issues/569}} 
The addition of document context to NER using BERT is discussed also by \newcite{virtanen2019multilingual}, who fill each input sample with the following sentences and use the first sentence in each sample for predictions, and thus only introduce context appearing \emph{after} the sentence of interest in the source text.

Of the related work discussed above, our approach most closely resembles that of \newcite{virtanen2019multilingual}, which in turn aims to directly follow \newcite{devlin2018bert}.
By contrast to other studies discussed above, we do not introduce extra features or embeddings representing cross-sentence information or incorporate extra information in addition to that captured by the BERT model. Instead, we directly utilize the BERT architecture and rely on self-attention and voting to combine predictions for sentences in different contexts.

\section{Data}

The data used in this study consists of pre-trained BERT models and NER datasets for five different languages. We aimed to use monolingual BERT models as numerous recent studies have suggested that well-constructed language-specific models outperform multilingual ones \cite{virtanen2019multilingual,vries_bertje_2019,le2020flaubert}.
We selected the following language-specific pre-trained BERT models for our study, focusing on languages that also have established benchmark data for NER:
\begin{itemize}
\item BERTje base, Cased for Dutch \cite{vries_bertje_2019}\footnote{\url{https://github.com/wietsedv/bertje}}
\item BERT-Large, Cased (Whole Word Masking) for English \footnote{\url{https://github.com/google-research/bert}}
\item FinBERT base, Cased for Finnish \cite{virtanen2019multilingual}\footnote{\url{https://github.com/TurkuNLP/FinBERT}}
\item German BERT, Cased for German \footnote{\url{https://deepset.ai/german-bert}}
\item BETO, Cased for Spanish \cite{CaneteCFP2020}\footnote{\url{https://github.com/dccuchile/beto}} .
\end{itemize}
For comparison purposes we also tested multilingual BERT\footnote{\url{https://github.com/google-research/bert}} with the Spanish language. From the models introduced above all except German and multilingual BERT have used the Whole Word Masking variation of the Masked Language Model objective in pre-training instead of the method introduced in the original paper \cite{devlin2018bert}. Whole Word Masking was introduced by the developers of BERT after the original paper was published. In this pre-training objective, all the tokens corresponding to one word in text are masked instead of completely random tokens, which often leaves some of the tokens in multi-token words unmasked.       
We aimed to apply sufficiently large, widely-used benchmark datasets for evaluating NER results, assessing our methods primarily on the CoNLL'02 and CoNLL'03 Shared task Named entity recognition datasets \cite{Tjong_Kim_Sang_2002,Tjong_Kim_Sang_2003}, which cover four of our five target languages. For the fifth language, Finnish, we use two recently published named entity recognition corpora \cite{ruokolainen2019finnish,luoma-EtAl:2020:LREC}\footnote{\url{https://github.com/mpsilfve/finer-data}}\textsuperscript{,}\footnote{ \url{https://github.com/TurkuNLP/turku-ner-corpus}}. These two Finnish datasets are annotated in a compatible way, and for this study they are combined into a single corpus by simple concatenation, following \newcite{luoma-EtAl:2020:LREC}.

\begin{table}[!t]
\centering
\begin{tabular}{l|lllll}
\textbf{Tokens}   & English &  German & Spanish & Dutch   & Finnish \\ \hline
Train             & 203,621 & 206,931 & 264,715 & 202,644 & 342,924 \\ 
Development       &  51,362 &  51,444 &  52,923 &  37,687 &  31,872 \\ 
Test              &  46,435 &  51,943 &  51,533 &  68,875 &  67,425 \\ 
\multicolumn{6}{c}{} \\
\textbf{Entities} & English &  German & Spanish & Dutch   & Finnish \\ \hline
Train             &  23,499 &  11,851 &  18,798 &  13,344 &  27,026 \\ 
Development       &   5,942 &   4,833 &   4,352 &   2,616 &   2,286 \\ 
Test              &   5,648 &   3,673 &   3,559 &   3,941 &   5,129 \\ 
\end{tabular}
\caption{Key statistics of the NER data sets}
\label{ner-data}
\end{table}

All of the NER datasets define separate training, development and test sets, and we follow the given subdivision for each. The training sets for each language are used for fine-tuning the corresponding BERT model for NER, development sets are used for evaluation in hyperparameter selection, and the test sets are only used in final experiments for evaluating models trained with the selected hyperparameters. As previous studies vary in whether to combine development data to training data for training a final model, we report also results where models are trained with a combined training and development set for final test experiments.
The datasets for the CoNLL shared task languages contain four different classes of named entities: Person (PER), Organization (ORG), Location (LOC) and Miscellaneous (MISC). The Finnish NER datasets also use the PER, ORG, and LOC types along with three others, Product (PROD), Event (EVENT), and Date (DATE). For implementation purposes we converted all the datasets to the same format prior to experiments: The character encoding of each file was converted to UTF-8, and the NER labelling scheme was converted to IOB2 \cite{ratnaparkhi1998maximum} also for corpora that were originally in the IOB scheme \cite{ramshaw-marcus-1995-text}. By contrast to the older IOB scheme, in the IOB2 scheme the label for the first token of a named entity is always marked with a B-prefix (e.g.\ B-PER), even if the previous token is not part of a named entity.
The key statistics for the NER datasets are presented in Table~\ref{ner-data}. Finally, we note that all the datasets except CoNLL'02 Spanish provide information on document boundaries using special \texttt{-DOCSTART-} tokens at the start of each new document.

\section{Methods}
\label{sec:methods}

As the starting point for exploring the cross-sentence information for NER using BERT, we use a NER pipeline implementation introduced by \newcite{virtanen2019multilingual} that closely follows the straightforward approach presented by \newcite{devlin2018bert}. Here, the last layer of the pre-trained BERT model is followed by a single time-distributed dense layer which is fine-tuned together with the pre-trained BERT model weights to produce the softmax probabilities of NER tags for input tokens. No modelling of tag transition probabilities or any additional processing to validate tag sequences is used.

\begin{figure}[!t]
\includegraphics[width=0.98\textwidth]{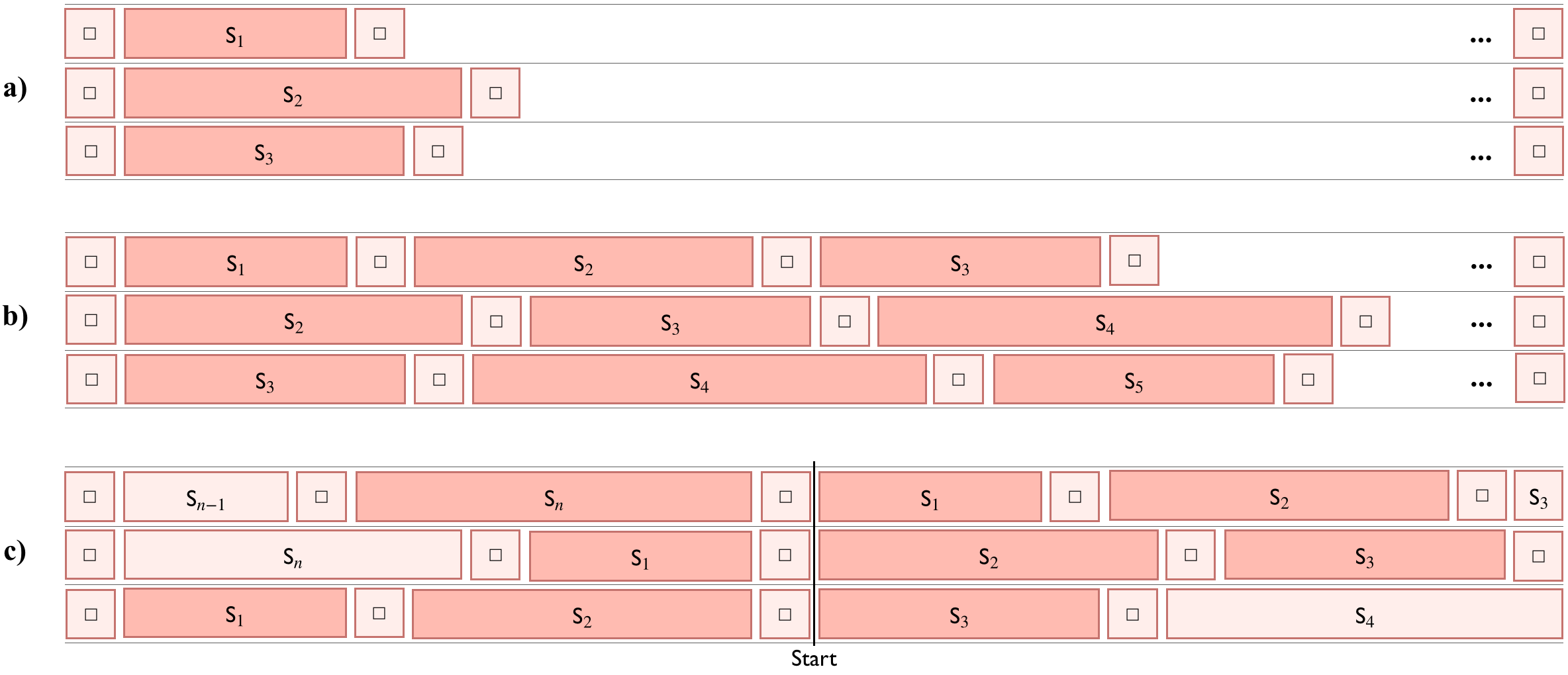}
\caption{Illustration of various input representations for sequence labelling tasks. a) One sentence per example (\emph{Single}), b) including following sentences (\emph{First, CMV}), c) including preceding and following sentences (\emph{Sentence in context}). CMV combines predictions for the same sentence (e.g. $\textrm{S}_2$ in b) in various positions and contexts. The empty square ({\small $\square$}) stands for special separator symbols (e.g.\ \texttt{[CLS]}, \texttt{[SEP]} and \texttt{[PAD]} for BERT); a light background color is used to represent special symbols and incomplete sentences in c).
}
\label{fig:context}
\end{figure}

In our implementation, exactly one example is constructed for each sentence of the corpus unless the sentence is so long that it does not fit to the maximum sequence length\footnote{In this special case the long sentence is split to produce multiple input sequences that are considered as sentences for the rest of the implementation.}. The sentence is placed at the beginning of the BERT window and following sentences from the corpus are used to fill the window (up to the maximum sequence length), with special separator (\texttt{[SEP]}) tokens separating the sentences. Partial sentences are used to fill up the BERT examples. As a special case, the sentences used for filling the window for the last sentences in input data are picked by wrapping back to the beginning of the corpus. This approach creates situations where some input samples contain sentences from different original documents, if the documents were next to one another in the corpus. For this reason, we also implemented documentwise wrapping of sentences if the input data had document boundaries marked with \texttt{-DOCSTART-} tokens. We used this information to build input samples by filling the sentences at the end of one document with the sentences from beginning of that same document instead of the next sentences in the original data. In this case only full sentences are added to each input sample, and padding (\texttt{[PAD]}) tokens are used to fill empty space if the next sentence in the input data does not fit into the window as demonstrated in (Figure~\ref{fig:context}b).

Constructing inputs in this way implies that the same sentences from the original data occur in different positions and with varying (sizes of) left and right contexts in different samples. We wanted to examine the predictions in different contexts more closely to see if there are consistent effects on tag prediction quality depending on the starting position of a sentence inside a context. One challenge here was how to consistently measure performance with different contexts: sentences are of different lengths, and as they are added to input samples, the beginning of the window was only place where the starting locations of sentences would align. Also, the number of sentences that fit into the window vary substantially. For this reason, it is not possible e.g.\ to always pick the N\emph{th} sentence to study as there are no guarantees one will exist in all examples. To address this issue and build input samples for testing predictions at different locations, we placed the sentence of interest to start at a specified location inside the window, and filled the window in both directions with sentences before / after the sentence of interest in the original data. We tested the starting positions of the sentence of interest from 1 (0 being the \texttt{[CLS]} token) up to the maximum sequence length (512 wordpieces) with intervals of 32 wordpieces. If the sentence of interest was longer than the space between a starting position and the maximum sequence length, the starting position for that particular sentence was moved backwards to fit the sentence in the window. 

Ensembles of classifiers are commonly used to improve classification performance at various tasks, and it seems reasonable to assume that predictions for the same input sentences in different positions and contexts create an ensemble-like construct. This is not an ensemble in the conventional sense, as the number of predictions we get for each sentence varies.
We evaluate two different variations combining the results from multiple predictions in different contexts. The first approach is to assign labels to sentences in each location first, and then take a majority vote of the assigned labels. The second approach is to add together the softmax probabilities of predictions in different contexts, and then take the argmax of the sum. For simplicity, we here term both Contextual Majority Voting (CMV) as they are variations of the same underlying idea. The implementation uses only predictions of tokens in whole sentences, not ones in partial sentences that may appear in input examples. 


For fine-tuning the pre-trained BERT models, we largely follow the process introduced in \cite{devlin2018bert}.
We use the maximum sequence length of 512 in all experiments to include maximal cross-sentence context, the Adam optimizer \cite{kingma2014adam} ($\beta_1 = 0.9$, $\beta_2 = 0.999$, $\epsilon = 1e-6$) with warmup of 10\% of samples, linear learning rate decay, a weight decay rate of 0.01, and norm clipping on 1.0. Sample weights are used for inputs so that the special tokens \texttt{[CLS]} and \texttt{[PAD]} are given zero weight and everything else 1 when calculating the loss (sparse categorical cross-entropy).
 
We select hyperparameters with an exhaustive search of the grid proposed by Devlin et~al., modified to skip batch size 32 and add batch sizes 2 and 4 instead as our initial experiments indicated better performance with smaller batch sizes. That is, the grid search is done over the following parameter ranges: 
\begin{itemize}
    \setlength\itemsep{-0.5em}
    \item Learning rate: 2e-5, 3e-5, 5e-5 
    \item Batch size: 2, 4, 8, 16
    \item Epochs: 1, 2, 3, 4
\end{itemize}
We repeated each experiment 5 times with every hyperparameter combination. The best hyperparameters were selected based on the mean of exact mention-level F1 scores, as evaluated against the development set using a Python implementation of the standard conlleval evaluation script.

As a reference we use a BERT model which is fine-tuned using only single sentences from the input data. For this baseline, predictions are also made on the basis of single sentences (see Figure~\ref{fig:context}a).

\begin{figure}[!t]
\begin{subfigure}{0.33\textwidth}
\includegraphics[width=\linewidth,height=4cm]{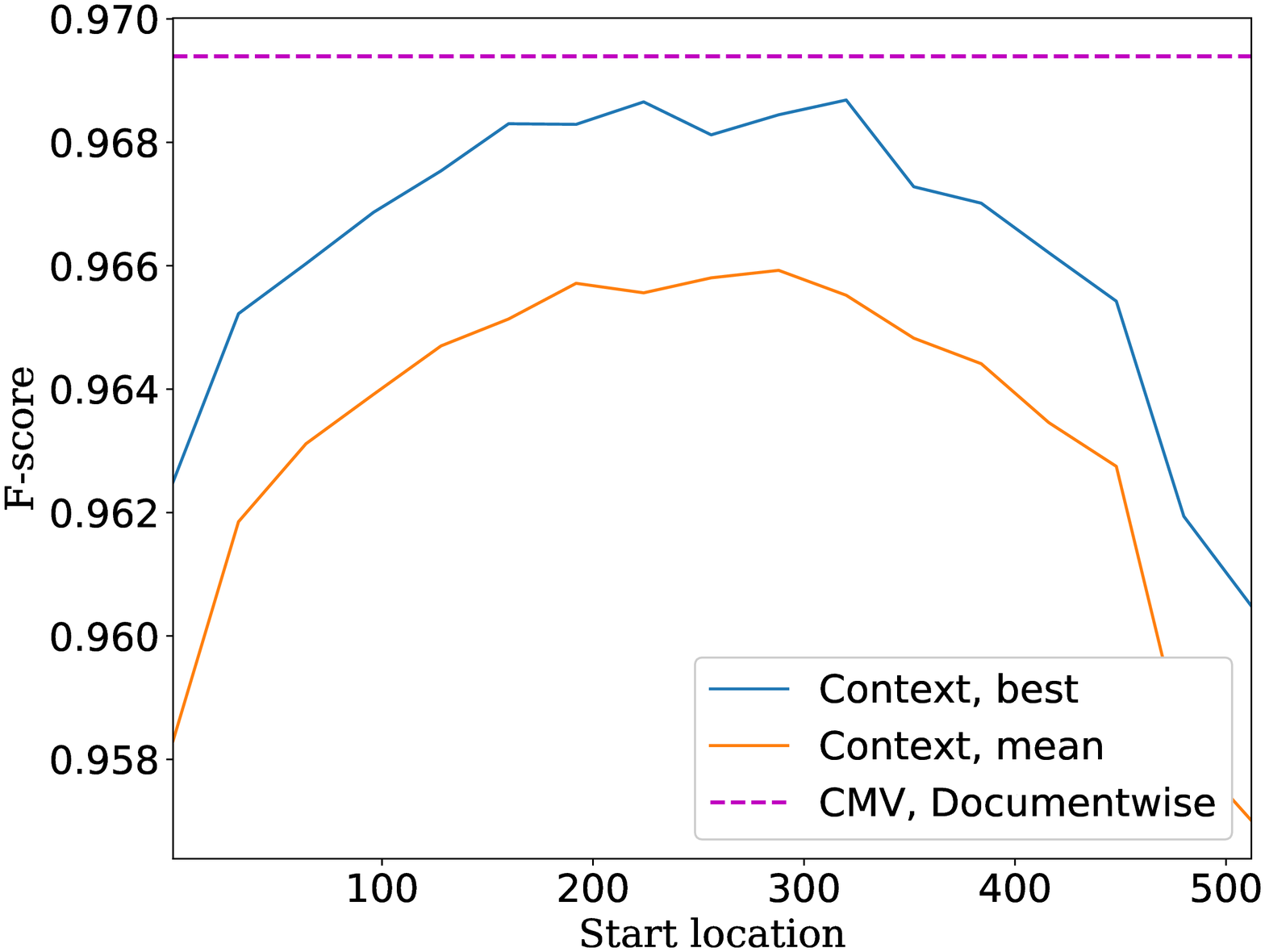} 
\caption{English}
\label{fig:subim1}
\end{subfigure}
\begin{subfigure}{0.33\textwidth}
\includegraphics[width=\linewidth, height=4cm]{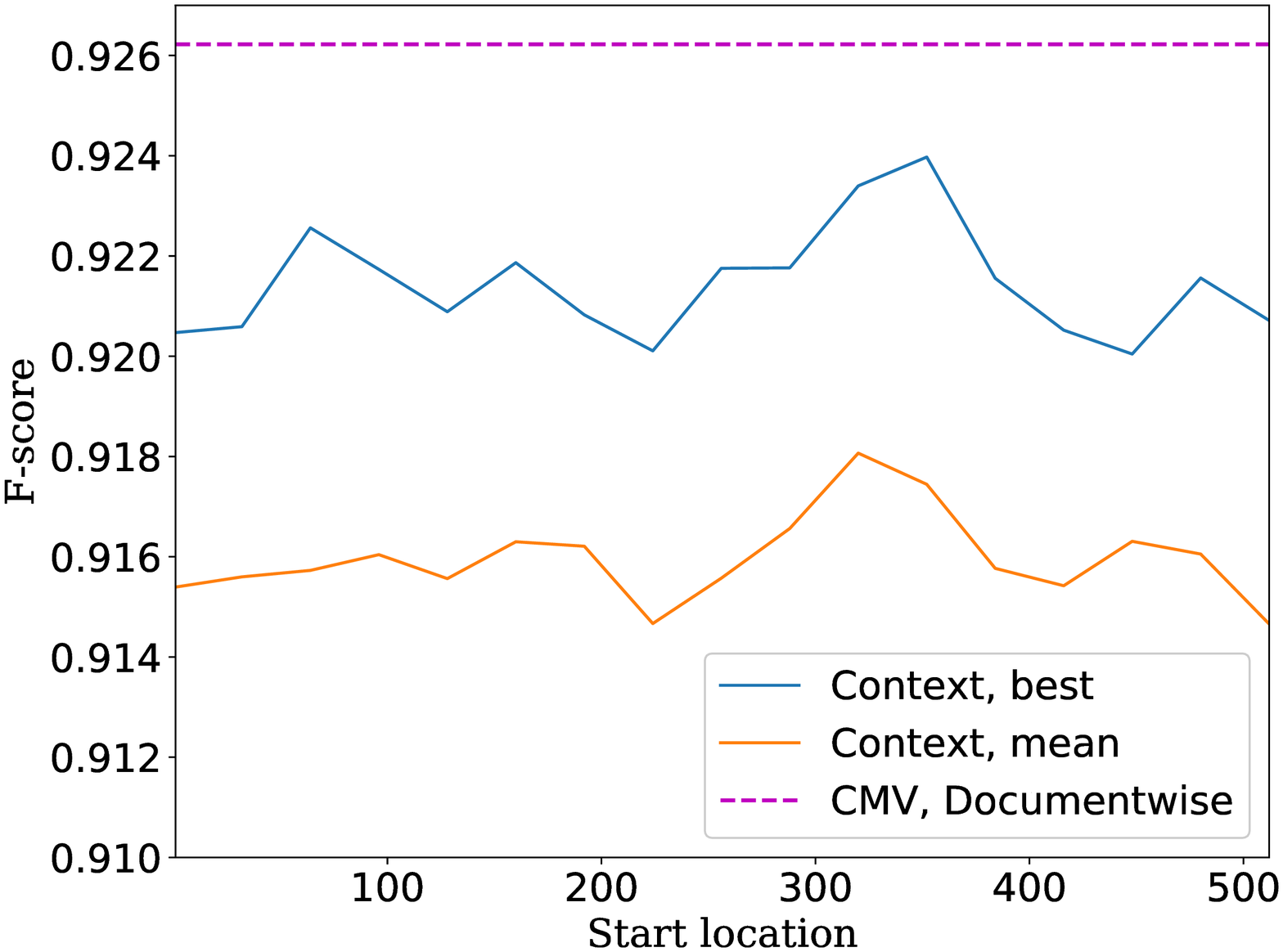}
\caption{Dutch}
\label{fig:subim2}
\end{subfigure}
\begin{subfigure}{0.33\textwidth}
\includegraphics[width=\linewidth, height=4cm]{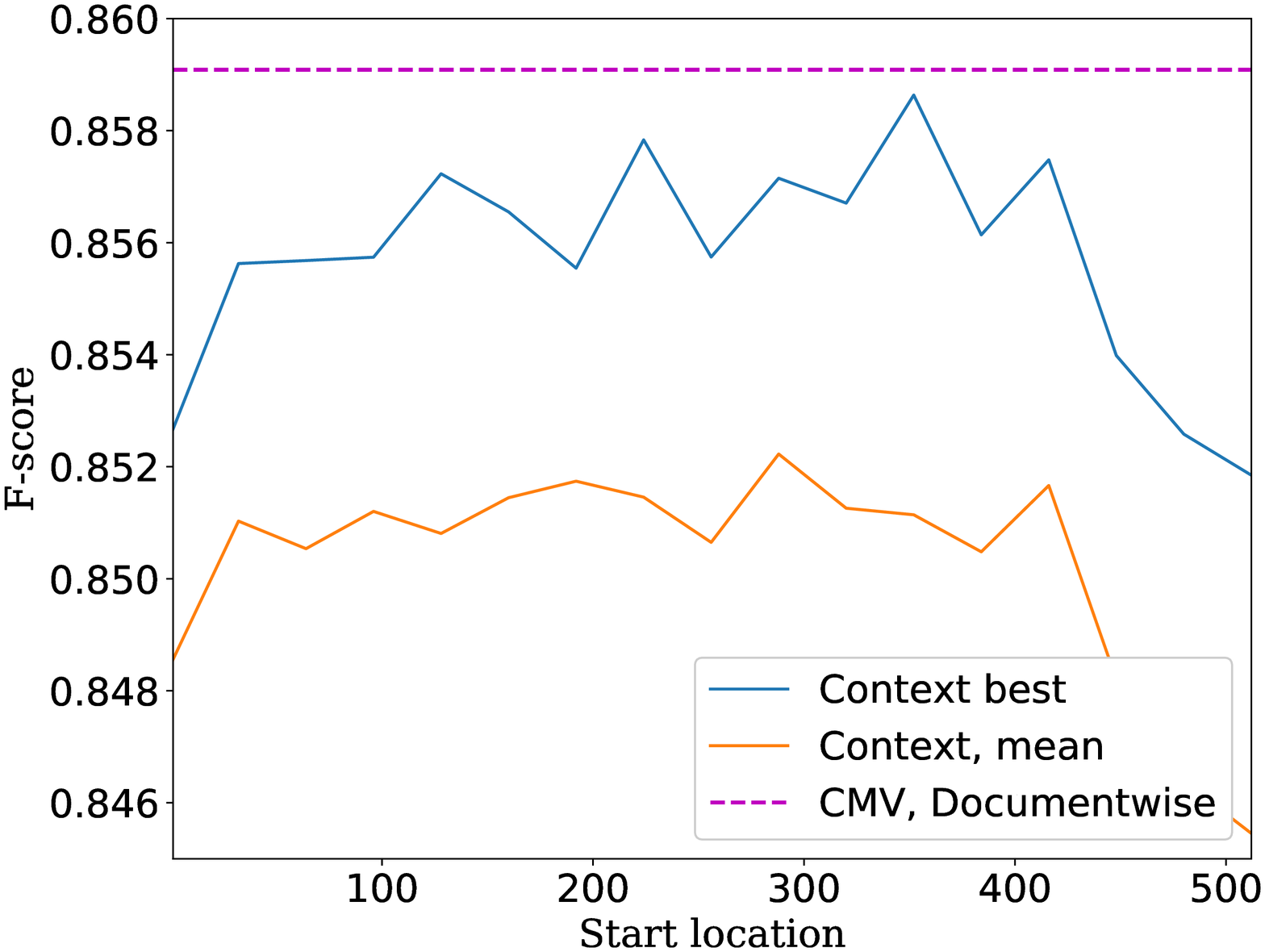} 
\caption{German}
\label{fig:subim3}
\end{subfigure}
\begin{subfigure}{0.33\textwidth}
\includegraphics[width=\linewidth, height=4cm]{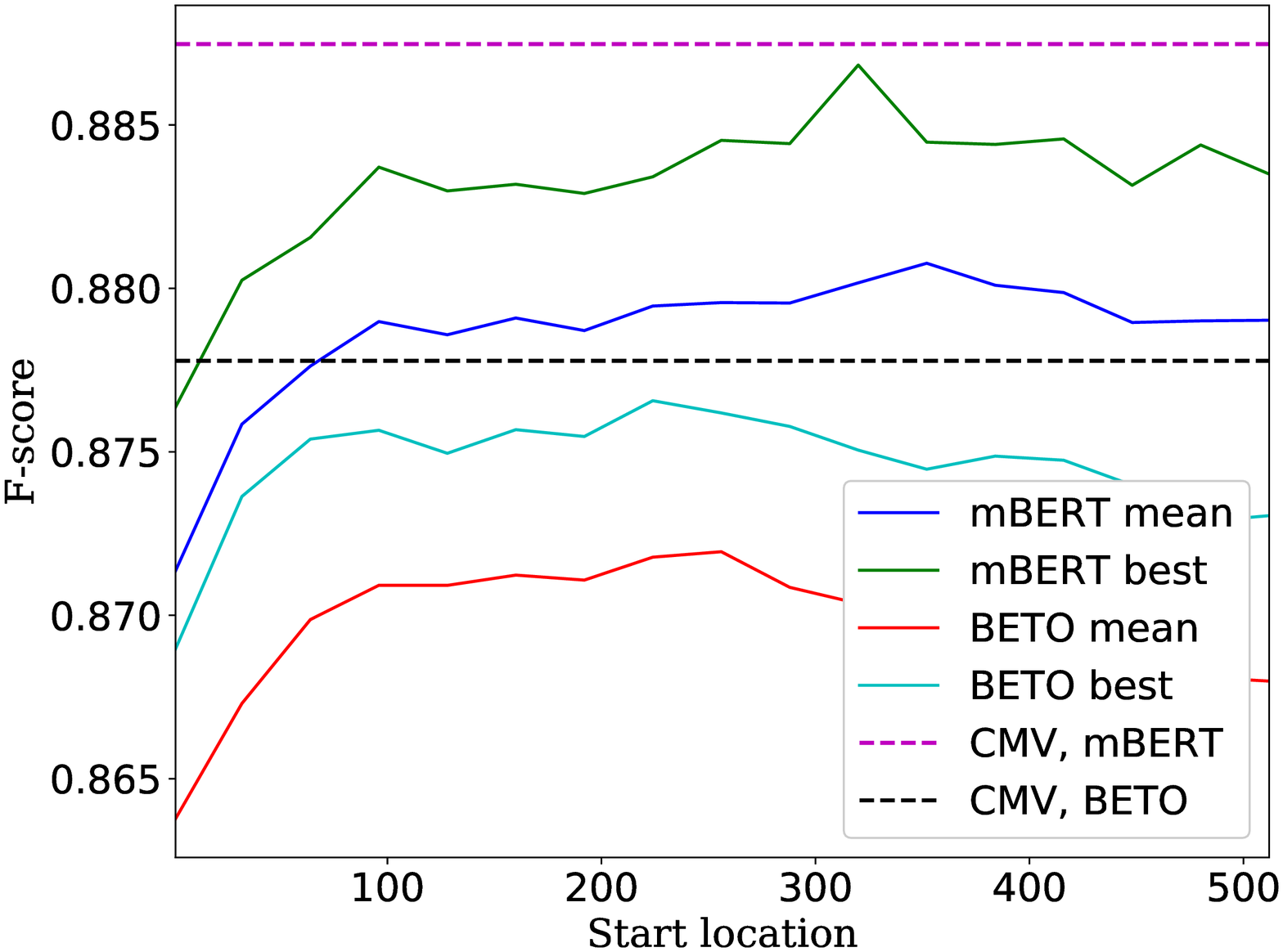}
\caption{Spanish}
\label{fig:subim4}
\end{subfigure}
\begin{subfigure}{0.33\textwidth}
\includegraphics[width=\linewidth, height=4cm]{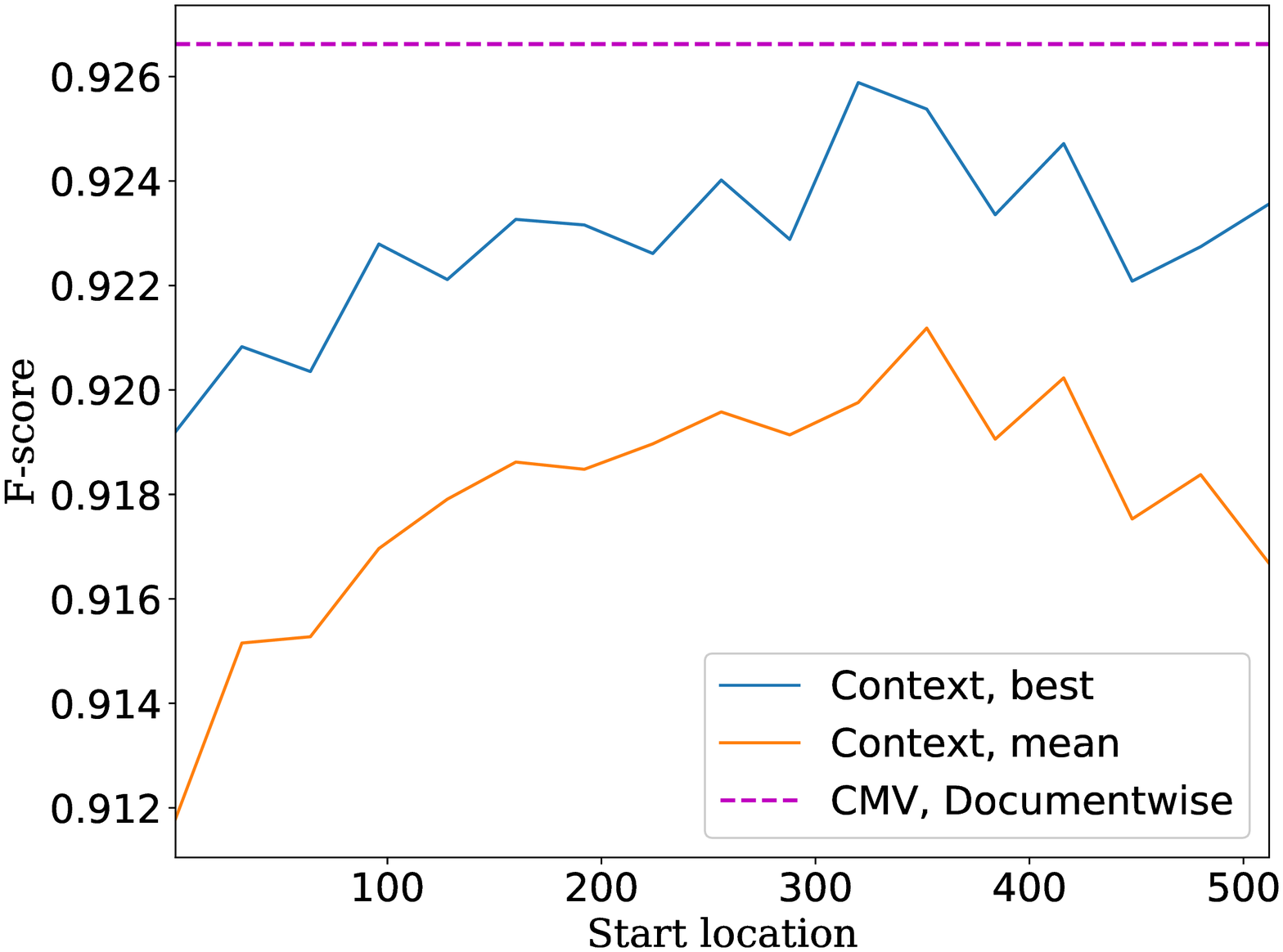}
\caption{Finnish}
\label{fig:subim5}
\end{subfigure}
\caption{NER performance on development set measured with CMV and in different sentence starting locations. The lower curves show mean performance over whole hyperparameter range, and the upper curves the results with the best hyperparameters (mean of 5 repetitions) for each location. The flat dashed lines show the best CMV results.}
\label{fig:image2}
\end{figure}

\section{Results}

Based on initial development set results, we decided to focus only on CMV using examples constructed document-wise of the variations of this method (see Section~\ref{sec:methods}). The exception here is the Spanish CoNLL dataset, for which document boundary information was not available. Further, as the differences between CMV variations were found not to be large, we decided to only consider the variant that first assigns labels and then votes between the labels. 

The effect of the sentence of interest starting location and the effect of CMV method on development data is illustrated in Figure~\ref{fig:image2}. Our initial expectation was that placing the sentence of interest near the middle of the sequence would generally yield the best performance. However, while this effect can be observed e.g.\ for English (Figure~\ref{fig:image2}a), the pattern does not hold in all cases, although in most cases performance does improve when moving the starting position away from either end of the context window. The problem was that the performance in the middle of the context did not appear to be stable enough to pick a reliable starting position to look at prediction time. This can be seen in the figure \ref{fig:image2} where the results for different starting locations tend to vary without a clear central optimum. The results for Dutch (Figure \ref{fig:subim2}) deviated the most from our expectations, and a possible reason for this was later found from the source data: the sentence order of the documents inside the original Dutch language data set has been randomized for copyright reasons. To test if randomizing the sentence order of documents has an effect on results, we tested this with other languages. However, in our initial experiments randomizing sentences inside each document did not result in significant performance drop on any of the tested languages. 

\begin{table*}[t!]
\centering
\begin{tabular}{ll@{\hskip 5pt}cl@{\hskip 5pt}cl@{\hskip 5pt}cl@{\hskip 5pt}c}
                            & \multicolumn{2}{c}{Precision} & \multicolumn{2}{c}{Recall} & \multicolumn{2}{c}{F1} & \multicolumn{2}{c}{F1 train+dev} \\ \hline
English, CMV             & 93.06 & (0.25) & \textbf{93.78} & (0.08) & 93.42 & (0.12) & 93.57 & (0.33) \\  
English, First          & \textbf{93.15} & (0.15) & 93.73 & (0.04) & \textbf{93.44} & (0.06) & \textbf{93.74} & (0.25)\\ 
English, Single         & 91.12 & (0.25) & 92.28 & (0.23) & 91.70 & (0.24) & 91.94 & (0.15)\\ \hline 
Dutch, CMV               & \textbf{93.12} & (0.26) & 93.26 & (0.18) & 93.19 & (0.21) &  \textbf{93.49} &  (0.23) \\ 
Dutch, First            & 93.03 & (0.65) & \textbf{93.38} & (0.38) & \textbf{93.21} & (0.51) & 93.39 & (0.26)\\ 
Dutch, Single           & 91.57 & (0.35) & 91.49 & (0.41) & 91.53 & (0.37) & 91.92 & (0.30)\\ \hline 
Finnish, CMV             & \textbf{92.91} & (0.18) & \textbf{94.42} & (0.13) & \textbf{93.66} & (0.13) & \textbf{93.78} & (0.26)\\ 
Finnish, First          & 92.56 & (0.14) & 94.24 & (0.08) & 93.39 & (0.10) & 93.65 & (0.26) \\ 
Finnish, Single         & 90.74 & (0.10) & 92.11 & (0.24) & 91.42 & (0.16) & 91.97 & (0.21) \\ \hline 
German, CMV              & \textbf{86.91} & (0.31) & \textbf{84.38} & (0.32) & \textbf{85.63} & (0.30) & \textbf{87.31} & (0.27)\\ 
German, First           & 86.37 & (0.39) & 84.07 & (0.10) & 85.21 & (0.22) & 86.91 & (0.11)\\ 
German, Single          & 85.55 & (0.20) & 81.81 & (0.31) & 83.64 & (0.21) & 85.67 & (0.25)\\ \hline 
Spanish, CMV              & \textbf{87.80} & (0.25) & \textbf{87.98} & (0.18) & \textbf{87.89} & (0.21) & \textbf{87.97} & (0.21)\\ 
Spanish, First           & 86.71 & (0.31) & 87.41 & (0.28) & 87.06 & (0.28) & 87.27 & (0.25)\\ 
Spanish, Single         & 87.43 & (0.53) & 87.90 & (0.34) & 87.66 & (0.43) & 87.52 & (0.41)\\ \hline 
S-mBERT, CMV              & \textbf{87.25} & (0.50) & \textbf{88.67} & (0.46) & \textbf{87.95} & (0.47) & \textbf{88.32} & (0.26)\\ 
S-mBERT, First          & 86.92 & (0.40) & 87.88 & (0.44) & 87.40  & (0.42) & 87.54 & (0.25) \\ 
S-mBERT, Single         & 87.19 & (0.28) & 87.81 & (0.26) & 87.50 & (0.26) & 87.57 & (0.29) \\ \hline 
\end{tabular}
\caption{NER results for different methods and languages (standard deviation in parentheses).}
\label{ner-results-language}
\end{table*}

The final test set results for models trained with the best hyperparameter combinations found using the development sets are summarized in Table~\ref{ner-results-language}. We report precision, recall and F1-score for models trained only on the training dataset, and additionally F1-scores for models trained with combined training and development sets using the same hyperparameters. For each language/BERT model pair, we report performance for the baseline using only a single sentence per window (Single), the approach where sentences from the following context are included but only predictions for the first sentence in each window are used (First), and, finally, performance with CMV (see also Figure~\ref{fig:context}).

These results show that BERT NER predictions systematically benefit from access to cross-sentence context. For all tested languages except Spanish, models that are fine-tuned and tested with samples containing context outperform models which do not use any context and are relying only on single sentences. What is not directly seen from Table~\ref{ner-results-language} is that generally the results with the method First outperform the results with the method Single, and similarly the method CMV generally outperforms the method First. Both English and Dutch seem to perform well with the method First and for Spanish the method Single also performs well. One thing to note is that English and Dutch results with CMV outperform the method First with the hyperparameters that produced the best result for the method First.
However, the final results for CMV just were not as good with the hyperparameters that produced the best performance for CMV on the development data. 

To further evaluate the performance of CMV method, we checked the results of each fine-tuned model on the development set during hyperparameter search. There were 48 hyperparameter combinations to evaluate for each model. For English, German, Spanish and Finnish, the CMV method outperformed the method First for every hyperparameter combination when calculating the results as the mean of mention-level F1 scores from 5 repetitions. For Spanish this includes both the experiments with the Spanish monolingual model as well as the experiments with the multilingual model. The only exception to this were the results on Dutch, for which CMV outperformed the method First in 41 cases out of 48. The fact that sentences in Dutch data are in randomized order may contribute to this. In total, the CMV method improved the results over method First in 281 cases out of 288.
In the same fashion, we evaluated the difference in performance between the method Single and the method First evaluated against the development set. The method First outperformed the method Single for every hyperparameter combination for every tested language.

\begin{table}[t!]
\small
\centering
\begin{tabular}{l|llll}
Model         & Our F1 &  Our F1 (t+d) & Current BERT & Current SOTA    \\ \hline
English   & 93.44 & \textbf{93.74} & 93.47 \cite{Liu_2019} & 93.5 \cite{Baevski_2019}\\ \hline
Dutch   & 93.21 & \textbf{93.49} & 90.94 \cite{Wu_2019} & 92.69 \cite{strakova-etal-2019-neural}\\ \hline
Finnish        &  93.66 &  \textbf{93.78} & 93.11 \cite{luoma-EtAl:2020:LREC} & 93.11 \cite{luoma-EtAl:2020:LREC} \\ \hline
German  &  85.63 & 87.31 & 82.82 \cite{Wu_2019} & \textbf{88.32} \cite{akbik2018coling} \\ \hline
Spanish   & 87.89 & 87.97 & 88.43 \cite{CaneteCFP2020} & \textbf{89.72} \cite{conneau-etal-2020-unsupervised} \\ \hline
Spanish, mBERT   & 87.95 & 88.32 & 88.43 \cite{CaneteCFP2020} & \textbf{89.72} \cite{conneau-etal-2020-unsupervised} \\ \hline
\end{tabular}
\caption{NER result comparison to the state of the art.}
\label{ner-sota2}
\end{table}

In Table \ref{ner-sota2} we compare the results using cross-sentence context with current the state-of-the-art in NER for the languages studied here. We are able to establish a new state-of-the-art result for three languages, English, Dutch and Finnish, as well as improve the best BERT-based score on German. These results benefit from using the combined training and development set in final model training. The previous state-of-the-art is also surpassed on Dutch and Finnish when only the training set is used for the final model. On Spanish our results fall slightly below the reported state-of-the-art. Perhaps somewhat surprising was that multilingual BERT outperformed the dedicated Spanish language BERT model, failing to replicate the results of \newcite{CaneteCFP2020}, who reported that the Spanish model outperformed that of \newcite{Wu_2019}, who had previously reached the best Spanish BERT performance using multilingual BERT.
Despite this minor discrepancy, we find that both the simple approach of including following sentences as context as well as CMV are very effective, allowing a straightforward BERT NER model to achieve state-of-the-art performance with only a few modifications of the representation.

\section{Discussion}

The results presented here are, as far as we know, the first systematic study on how cross-sentence information can be utilized with BERT for NER, and the methods presented here form a good starting point for discussion and further research into the subject. Contextual Majority Voting is straightforward to implement in existing BERT-based systems as the actual model and associated infrastructure is not modified. It is quite probable that similar ways of including cross-sentence information or majority voting structures may be beneficial with other attention-based models as well. The computational overhead for the required pre- and postprocessing of the samples is very modest, but increasing the maximum sequence length in fine-tuning e.g. from 128 to 512 to fit more sentences in one sample does come with a tradeoff of increased computational cost.

One aspect deserving more study is how prediction performance is affected if sentences are not repeated, or repeated fewer times, in examples during prediction. Reducing or entirely avoiding repetition would allow for more efficient use of the model while still providing context for sentences, which might be a reasonable compromise between performance and computational efficiency for large-scale practical applications.
A further possibility for future research would be to explore weighted majority voting. Our results lend some support to the idea that predictions made for tokens around in the center of the window are generally more reliable than predictions for tokens near its edges, where context is limited on one side of the token. Providing higher weight to predictions in the middle of the sequence could potentially help further improve the performance of the aggregation approach. Another aspect for future work would be to study the effect of the context and sentence order. Our preliminary tests with randomized sentence order from same documents showed minimal effect on performance. Is it enough to have context from the same document? Would the situation change drastically if random sentences from the whole input data were used instead? Finally, the incorporation of transition probabilities or other processing to check tag sequences for illegal transitions would likely improve performance further.

\section{Conclusions}

We have presented a comprehensive evaluation of the effect of including cross-sentence context for named entity recognition with BERT and introduced a simple and easy-to-implement approach for the task using majority voting. The proposed method established new state-of-the-art results in named entity recognition for three languages and is near the state-of-the-art for two other languages, demonstrating how simple ideas may boost the performance of even very strong models.

We release all methods implemented in this work under open licenses from \url{https://github.com/jouniluoma/bert-ner-cmv} .

\section*{Acknowledgements}

We wish to thank the CSC -- IT Center for Science, Finland, for generous computational resources. This work was funded in part by the Academy of Finland.

\bibliographystyle{coling}
\bibliography{coling2020}

\end{document}